\documentclass[10pt,twocolumn,letterpaper]{article}

\usepackage{iccv}
\usepackage{times}
\usepackage{epsfig}
\usepackage{graphicx}

\usepackage{tikz}
\usepackage{comment}
\usepackage{amsmath,amssymb} 
\usepackage{amsfonts}
\usepackage{color}
\usepackage{algorithm}
\usepackage{algorithmic}
\usepackage{multicol}
\usepackage{lipsum}
\usepackage{mwe}
\usepackage{caption}
\usepackage{subcaption}
\usepackage{array}
\usepackage{multirow}
\usepackage{multicol}
\usepackage{makecell}
\usepackage{booktabs}
\usepackage{wrapfig}
\usepackage[resetlabels]{multibib}
\newcites{app}{References}

\DeclareMathOperator*{\argmax}{arg\,max}

\newtheorem{assumption}{Assumption}

\usepackage[breaklinks=true,bookmarks=false]{hyperref}

\iccvfinalcopy 

\def\iccvPaperID{****} 

\ificcvfinal\fi

\begin{document}

\title{OpenIncrement: A Unified Framework for Open Set Recognition and Deep Class-Incremental Learning}

\author{Jiawen Xu\\{\small Technical University Berlin}\\
{\tt\small jiawen.xu@campus.tu-berlin.de}
\and
Claas Grohnfeldt \\ {\small Huawei Munich Research Center} \\
{\tt\small claas.grohnfeldt@huawei.com}
\and
Odej Kao\\
{\small Technical University Berlin}\\
{\tt\small odej.kao@tu-berlin.de}
}

\maketitle
\ificcvfinal\thispagestyle{empty}\fi

\begin{abstract}
   In most works on deep incremental learning research, it is assumed that novel samples are pre-identified for neural network retraining. However, practical deep classifiers often misidentify these samples, leading to erroneous predictions. Such misclassifications can degrade model performance. Techniques like open set recognition offer a means to detect these novel samples, representing a significant area in the machine learning domain.

    In this paper, we introduce a deep class-incremental learning framework integrated with open set recognition. Our approach refines class-incrementally learned features to adapt them for distance-based open set recognition. Experimental results validate that our method outperforms state-of-the-art incremental learning techniques and exhibits superior performance in open set recognition compared to baseline methods.
\end{abstract}

\section{Introduction}  \label{sec-introduction}

Deep learning-based classification, underpinning advancements in areas like object recognition \cite{zhao2019object} and sentiment analysis \cite{do2019deep}, heavily relies on ample pre-collected data. In realistic scenarios, accumulating extensive training data quickly is often infeasible due to episodic and unpredictable data emergence. Thus, the appeal of deep incremental learning, which assimilates new data while retaining prior knowledge \cite{rebuffi2017icarl, castro2018end, wu2019large, zhao2020maintaining}. This mode of learning eschews retraining from scratch, eliminating the need for complete datasets during successive \emph{training sessions}. Among various continual learning paradigms, class-incremental learning, wherein models assimilate new classes and retain information about observed ones, is paramount.

A notable oversight in the realm of class-incremental learning is the prerequisite identification of novel samples or \emph{open sets}. While numerous works have addressed the \emph{Open Set Recognition} (OSR) challenge \cite{bendale2016towards, Dhamija2018ReducingNA, hassen2020learning, jia2021mmf, miller2021class, sun2021m2iosr, Saito2021OpenMatchOC}, only a handful have sought an integrative framework combining OSR and incremental learning. While \cite{joseph2021open} charted this terrain within object detection, their approach was restrictive in scope and application. It's imperative to establish a singular model adept at OSR that also evolves with new classes.

Our investigation aligns closely with task-free continual learning \cite{aljundi2019task, jin2021gradient, ye2022task}, a domain where the onus is on innate model mechanisms to discern data drifts and instigate model updates. Notably, these works diverge in their handling of new class identification and rarely evaluate this performance aspect. Our proposition centers on a unified model employing distillation-based continual learning.

In this work, we meld class-incremental learning with OSR considerations, leaning on distance-based OSR and rehearsal-infused continual learning (see \ref{subsec-OSR} and \ref{subsec-incremental-learning}). The central challenge is maintaining discernible inlier and outlier features post-training. We discern that feature spaces undergo \emph{distortions} during incremental training, leading to intertwined inlier and outlier features—a probable cause of the \emph{catastrophic forgetting} phenomenon.

Response-based knowledge distillation \cite{hinton2015distilling}, pivotal in staving off \emph{catastrophic forgetting} in deep incremental learning, often bypasses feature relation transfer, resulting in distortions. We counter this by leveraging relation-based knowledge distillation (RKD) and supplement with supervised contrastive learning (\emph{SupCon}) \cite{khosla2020supervised} to bolster class-specific feature separation. Unlike the method in \cite{cha2021co2l}, our focus is a holistic framework for OSR and class-incremental learning, not merely an incremental learning strategy. Future explorations might traverse other OSR and incremental learning intersections.

Our primary contributions are:
\begin{itemize}
    \item Pinpointing feature distortion as an aftermath of incremental learning.
    \item Crafting a cohesive framework for open set recognition and class-incremental learning.
    \item Empirical validation of our approach on public and continual learning benchmarks, achieving commendable results for both facets.
\end{itemize}

\section{Background} \label{sec-background}
Since our work relates to both open set recognition and class-incremental learning, we will review the main approaches of these two domains in this section. Meanwhile, our method is closely connected to knowledge distillation. We will therefore briefly introduce it as well. 

\subsection{Open Set Recognition}    \label{subsec-OSR}
Open set recognition is an everlasting topic for supervised classifiers while the openness of the data is usually unknown and the novel classes can always appear and should be detected.
For deep neural networks, there are mainly three categories of OSR methods, namely distance-based \cite{bendale2016towards}\cite{Dhamija2018ReducingNA}\cite{hassen2020learning}\cite{jia2021mmf}\cite{miller2021class}\cite{sun2021m2iosr}\cite{yang2020convolutional}\cite{arpl}, generative model-based \cite{sun2020open}\cite{Cao2021OpenSetRW}\cite{perera2020generative}\cite{yoshihashi2019classification}, and background class-based \cite{hendrycks2019oe}\cite{Yu2017Adversarial}\cite{Ge2017GenerativeOpenMax}. 

Bendale et al. have first proposed \emph{OpenMax} that models the \emph{Euclidean} distances between the softmax outputs of each data sample and their closest class centers using \emph{Weibull} distribution \cite{bendale2016towards}. The outliers are then discriminated by thresholding the inference probability. Moreover, Dhamija et al. proposed the \emph{Objectosphere} loss \cite{Dhamija2018ReducingNA} and Hassen et al. presented their \emph{ii-loss} \cite{hassen2020learning} to learn features that are better clustered.
Miller et al. proposed the \emph{Class Anchor Clustering} loss that enforces the inlier features to be close to each other \cite{miller2021class}. There are more works introducing similar methods, such as \cite{sun2021m2iosr} and \cite{Saito2021OpenMatchOC}.

Generative model-based methods directly model the inliers. In \cite{sun2020open}, variational autoencoders are trained using inliers to recognize outliers. Analogously, Cao et al. applied \emph{variational autoencoders} to learn the latent features for inliers that are used for both inlier class clustering and outlier detection \cite{Cao2021OpenSetRW}. 
Perera et al. proposed to augment the inlier samples with GANs to gain more informative features that can be more discriminative from the outliers \cite{perera2020generative}. 

Background class-based methods are straightforward in that one extra outlier class output is added to the original classifiers and the entire models are trained in a supervised way. 
Yu et al. proposed to generate pseudo outliers using GANs inspired by the principle of \emph{adversarial training} \cite{Yu2017Adversarial}. The generated pseudo samples are then utilized altogether with the inliers to train a $(C+1)$ classifier ($C$ is the number of inlier classes). Analogously, Ge et al. trained a $(C+1)$ classifier with synthetic outliers to calibrate the feature distances in \emph{OpenMax} \cite{Ge2017GenerativeOpenMax}.
The drawback of these methods lies in that the open set classes are infinite and cannot be completely collected, especially the "\emph{unknown unknowns}".

\subsection{Class-Incremental Learning} \label{subsec-incremental-learning}
The challenge behind deep incremental learning is \emph{catastrophic forgetting} that the models will lose the capability for observed tasks after the new training session. 
To prevent catastrophic forgetting, there are two main categories of class-incremental learning approaches, namely  parameter-based methods \cite{kirkpatrick2017overcoming}\cite{zenke2017continual} and distillation-based methods \cite{rebuffi2017icarl}\cite{castro2018end}\cite{Hou_2019_CVPR}\cite{li2017learning}. 

Parameter-based methods require no stored history exemplars and apply constraints directly on the model parameters. \emph{Elastic Weight Consolidation} (\emph{EWC}) \cite{kirkpatrick2017overcoming} proposed to constrain the weights in the region of low-performance degradation for observed classes. \emph{Synaptic Intelligence} (\emph{SI}) \cite{zenke2017continual} regulates the weights during new training sessions to keep the important weights from significant changes. 

The distillation-based methods apply knowledge distillation (see \ref{subsec-knowledge-distillation}) in new training sessions to directly transfer the knowledge learned by old models (i.e. the teacher) to the new models (i.e. the student). A portion of observed data (i.e., exemplars) should be stored in these methods.
\emph{Learning without Forgetting} (\emph{LwF}) \cite{li2017learning} has first applied response-based knowledge distillation in incremental learning. \emph{Incremental classifier and Representation Learning} (\emph{iCaRL}) \cite{rebuffi2017icarl} adapts a similar knowledge distillation approach but classifies the test samples with \emph{nearest-mean-of-exemplars} approach instead of the final softmax layer because of the class imbalance. 

\subsection{Knowledge Distillation} \label{subsec-knowledge-distillation}
In the context of deep learning, knowledge distillation (KD) was first known in \cite{hinton2015distilling} and \cite{romero2014fitnets} that enables to transfer the knowledge learned by a larger model (teacher) to a smaller one (student). There are three types of knowledge that can be transferred, namely response-based knowledge, feature-based knowledge, and relation-based knowledge.  

Response-based knowledge distillation is to directly align the predictions between students and teachers. The  divergence  between the logit layers of teachers and students is encouraged to be minimized. 
In \cite{hinton2015distilling}, the logit layer of the teacher network is adapted as \emph{soft targets} when computing the softmax loss so that the student can be expected to give the same outputs as the teacher. 

In feature-based knowledge distillation, not only the last logit layer but also the intermediate layers of the teachers are utilized to transfer knowledge. The students can therefore output similar features as the teachers. In \cite{romero2014fitnets}, the neural activations of the first layers in the teacher model are directly used to match the student model. Kim et. al. proposed to extract the \emph{factor maps} of teachers' and students' layers using convolutional modules and let the student factor mimic the teacher factor \cite{kim2018paraphrasing}. Similarly, Passban et. al. proposed a combinatorial technique that can merge the features of multiple layers in the teacher model using the attention mechanism \cite{passban2020alp}. 

Unlike the previous two categories that align the layer outputs between the teachers and students, relation-based knowledge distillation tends to preserve the relationships among the features in different layers or instances \cite{park2019relational}\cite{tung2019similarity}\cite{tian2019contrastive}. Park et. al. proposed the relational knowledge distillation that transfers the instance relations between teachers and students \cite{park2019relational}. Tung et. al. proposed the similarity-preserving knowledge distillation, in which the similarities between the instance features can be transferred to the students \cite{tung2019similarity}.

\section{Problem Statement} \label{sec-problem-statement}
The goal of this study is to enable neural networks to learn representations that are suitable for open set recognition under class-incremental learning settings. Distance-based OSR methods are often based on two assumptions (see Assumption ~\ref{assumption-intra} and ~\ref{assumption-inter} below) \cite{miller2021class}\cite{hassen2020learning}\cite{yang2020convolutional}\cite{arpl}, which are from the \emph{Smoothness Assumption} \cite{cohen2003numerical} stating that \emph{if two points are near, their corresponding output values cannot be arbitrarily far from each other}.
The problem is then converted to how to enable the learned representations to still hold these two assumptions after repeating class-incremental training. 

\begin{assumption}
Data representations of the same class should be close to each other.
\label{assumption-intra}
\end{assumption}

\begin{assumption}
Data representations of different classes should be pushed apart.
\label{assumption-inter}
\end{assumption}

The metrics \emph{intra spread}, $S_{intra}$, in Equ. \eqref{equ-intra-class}, and \emph{inter spread}, $S_{inter}$, in Equ. \eqref{equ-inter-class} are introduced in \cite{hassen2020learning} to evaluate how well the learned representations follow the above two assumptions respectively. In Equ. \eqref{equ-intra-class} and \eqref{equ-inter-class} and the following text, $\mathbf{\mu}$ represents the euclidean centers of each class and $\mathbf{z}_i$ are the instance representations. The number of inlier classes is denoted using $K$, and $C_j$ is the number of instances of each class.
According to Assumption \ref{assumption-intra} and \ref{assumption-inter}, $S_{inter}$ should be as large as possible whereas $S_{intra}$ is supposed to be small. However, the distances are of different scales in different models and are hard to be compared directly.
We, therefore, propose to fuse these two metrics using their ratio, i.e., $R_s = \frac{S_{intra}}{S_{inter}}$. It is straightforward to understand that the smaller the $R_s$ is, the more suitable the feature is for OSR. 

\begin{equation}
    S_{intra} = \frac{1}{N} \sum_{j=1}^{K} \sum_{i=1}^{C_j} {\| \Vec{\mathbf{\mu}_i} - \Vec{\mathbf{z}_i}\|}_2^2
\label{equ-intra-class}
\end{equation}

\begin{equation}
    S_{inter} = \min_{\substack{{1 \leq m  \leq K} \\{m+1 \leq n  \leq K}}}{\| \Vec{\mathbf{\mu}_m} - \Vec{\mathbf{\mu}_n}\|}_2^2
\label{equ-inter-class}
\end{equation}

\begin{table*}[h]
\centering
\begin{tabular*}{\textwidth}{c @{\extracolsep{\fill}} ccccccccccc}
 \toprule
 \#Classes & 10 (base) & 20  & 30 & 40 & 50 & 60 & 70 & 80 & 90 & 100  \\ [0.5ex] 
 \hline
 Joint & 2.54  & 3.38 & 3.76 & 3.95 & 4.06 & 3.98 & 4.07 & 3.65 & 3.85 & 3.82\\ 

 SI \cite{zenke2017continual} & 2.54  & 5.56 & 6.64 & 6.61   & 6.82 & 7.42  & 7.33 & 7.25 & 9.83 & 7.25  \\ 
 EWC \cite{kirkpatrick2017overcoming} & 2.54  & 11.70 & 10.14 & 9.83 & 10.64 & 13.48  & 14.96  & 15.17 & 13.79 & 12.28 \\ 

 iCaRL  \cite{rebuffi2017icarl} & 2.54 & 3.41 & 3.40 & 6.87 & 5.70 & 5.98 & 6.06 & 6.11 & 7.05 & 7.15 \\ 

 LwF  \cite{li2017learning} &2.54  & 5.11 & 4.74 & 6.41 & 6.66 & 6.05 & 6.29 & 5.52 & 6.37 & 7.44 \\ 

 \bottomrule
\end{tabular*}
\caption{$R_s$ for joint retraining and incremental training using methods in \cite{zenke2017continual}\cite{kirkpatrick2017overcoming}\cite{rebuffi2017icarl}\cite{li2017learning} on \emph{Cifar-100} dataset \cite{cifar}. The dataset is split into 10 tasks with 10 novel classes each. The base 10-class models are the same for all methods.
All models are trained using the best configurations given in \cite{buzzega2020dark}. }
 \label{table-rs}
\end{table*}

It can be found from Tab. \ref{table-rs} that $R_s$ in incremental learning settings are much larger than in normal training (line \emph{Joint} in the table). We believe it is because the feature relations between the data samples are not preserved when the models are updated incrementally. For incremental learning methods based on rehearsal and regularization, such as \emph{iCaRL}, the knowledge learned in previous sessions is preserved mainly through response-based knowledge distillation. As stated in \cite{park2019relational}, response-based knowledge distillation cannot transfer data-sample relations between teachers and students. Therefore, Assumption \ref{assumption-intra} and \ref{assumption-inter} can be easily broken with these incremental learning methods. 
For regularization-based incremental learning methods, such as \emph{EWC} and \emph{SI}, the critical weights for observed classes are constrained to change during retraining to prevent catastrophic forgetting. In these approaches, the newly optimized models have achieved balances between the loss functions of old and new tasks. But they ignored maintaining the relations between the features as well.  

In order to prevent such a phenomenon, the main research question in this study is to search for an incremental learning method, with which the above two assumptions for data representations can still be maintained after repeating incremental training. And the newly learned representations are hence adaptable for OSR. 

The problem we address in this paper can be formulated as follows: the labeled observed training data is denoted using $\mathcal{\mathbf{D}}_{obser}= (\mathbf{X}_{obser}, \mathbf{Y}_{obser}) =\{\mathbf{x}_i, y_i\}$ and the label $y_i \in \mathbf{C}_{obser} = [0, 1,..., c_{obser}-1]$. A deep classifier $\mathbf{F}(\mathbf{x}_i) \rightarrow \hat{y}_i$ is trained using $\mathcal{\mathbf{D}}_{obser}$. $\hat{y}_i$ denotes the predictions and $\mathbf{F}(\mathbf{x}_i) = (\mathbf{H}\circ\mathbf{E})(\mathbf{x}_i)$, in which $\mathbf{E}$ and $\mathbf{H}$ are the deep encoder and classification header respectively. $\mathbf{z}_i = \mathbf{E}(\mathbf{x}_i)$ represents the feature maps outputted by the deep encoder.
When the new classes $\mathbf{C}_{new}$ appear, then $\mathcal{\mathbf{D}}_{obser}=\mathcal{\mathbf{D}}_{obser} \cup \mathcal{\mathbf{D}}_{new}$, $c_{obser} = c_{obser} + c_{new}$ and $\mathbf{F}(\cdot), \mathbf{E}(\cdot)$ and $\mathbf{H}(\cdot)$ will be updated using incremental learning approach $\mathcal{I}$ so that $\mathbf{F'}(\cdot)=\mathcal{I}(\mathbf{F}(\cdot))$, $\mathbf{H'}(\cdot)=\mathcal{I}(\mathbf{H}(\cdot))$ and $\mathbf{E'}(\cdot)=\mathcal{I}(\mathbf{E}(\cdot))$. We intend to optimize $\mathcal{I}$ in order to let $R_s$ over $\mathbf{E'}(\mathbf{x}_{obser})$ be as small as possible.

\section{Methods}  \label{sec-methods}

This study is aimed to propose a deep class-incremental learning method in which the learned features can fulfill Assumption \ref{assumption-intra} and \ref{assumption-inter} for distance-based open set recognition. 
As discussed in Sec. \ref{sec-background} and \ref{sec-problem-statement}, deep incremental learning using response-based knowledge distillation loss cannot transfer the instance relations to the new models and therefore fails the OSR methods. Hence, we apply relation-based knowledge distillation in this research. Furthermore, compared with cross-entropy loss, which is the most applied training strategy in classification problems, supervised contrastive learning is capable of learning features that satisfy Assumption \ref{assumption-intra} and \ref{assumption-inter} better as we will prove in the following text. In summary, the deep incremental learning approach proposed in this study is based on supervised contrastive learning and relation-based knowledge distillation. We name it \textbf{\emph{OpenIncrement}}. The loss function is in Equ. \eqref{equ-loss-total}, in which $\mathcal{L}_{SupCon}$ stands for supervised contrastive loss and $\mathcal{L}_{distill}$ represents the distillation loss. $\alpha$ is the hyperparameter to balance both. 

\begin{figure*}
  \centering
  \includegraphics[
  width=1.0\textwidth,%
  ]{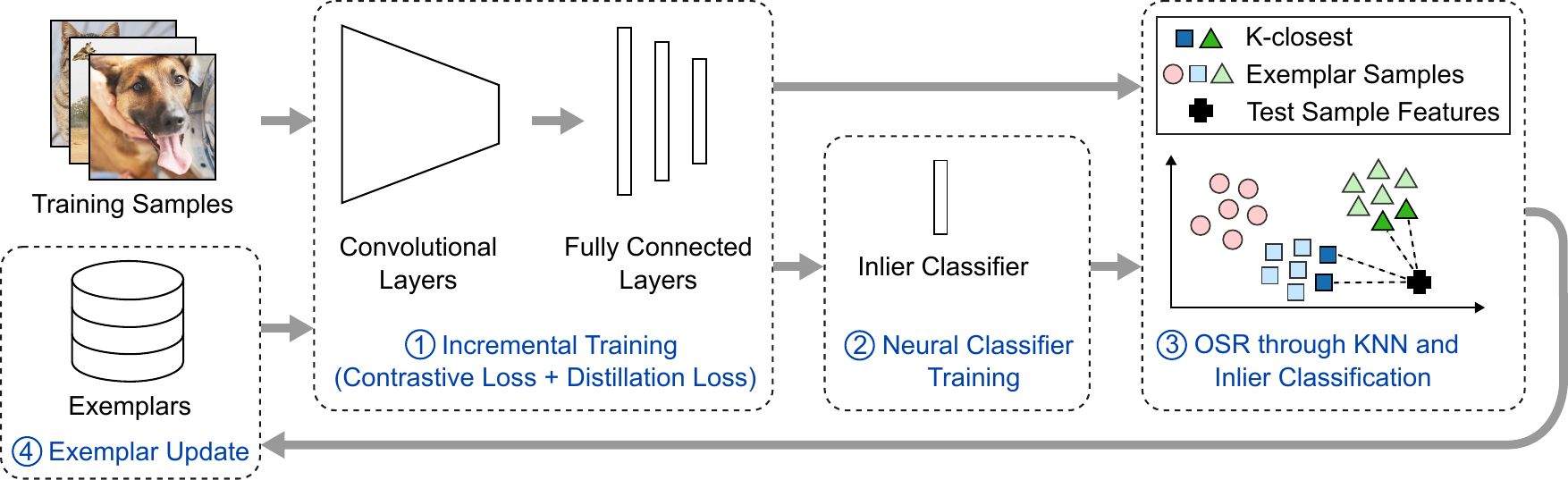}
  \caption{Illustration of our proposed framework. 1. A CNN encoder (backbone) is continuously trained with supervised contrastive learning and relation-based knowledge distillation. 2. A neural classifier after that is then trained with cross-entropy loss.  3. During testing, the features of the testing samples and exemplars given by the backbone are compared and outliers are detected using K nearest neighbor method. The inliers are classified using the neural classifier. 4. Exemplars are updated after each training session. Better viewed in color.}
  \label{fig-framework}
\end{figure*}
 
\begin{equation}
\mathcal{L}_{total} = \alpha*\mathcal{L}_{SupCon} + (1 - \alpha)*\mathcal{L}_{dis}
\label{equ-loss-total}
\end{equation}

\subsection{Supervised Contrastive Learning} \label{subsec-supcon}

In contrastive learning, the samples from the same class (positive sets) are pushed closer in feature space and vice versa. Normally, contrastive learning is applied in a self-supervised fashion, in which the positive sets are formed using data augmentation or co-occurrence. Khosla et al. developed contrastive learning to a supervised manner so that the labels can guide the selection of positive and negative sets and have gained higher accuracy in image recognition tasks than cross-entropy loss \cite{khosla2020supervised}. 
The supervised contrastive loss function is shown in Equ. \eqref{equ-contra-loss}, in which $\mathbf{z}_i$ and $\mathbf{z}_{p}$ are positive feature pairs belonging to positive set $P(i)$, and $A(i)$ stands for the negative sets for $\mathbf{z}_i$. $\tau$ is the temperature scaling factor that can tune the discrimination between the positive and negative sets. 

\begin{equation}
    \mathcal{L}_{SupCon} = - \sum_{i \in I} \frac{1}{|P(i)|} \sum_{p \in P(i)} \log \frac{\exp (\textbf{z}_i \cdot \textbf{z}_{p} / \tau)}{\sum_{a \in A(i)} \exp(\textbf{z}_i \cdot \textbf{z}_a / \tau)}
    \label{equ-contra-loss}
\end{equation}

\begin{figure}
  \begin{center}
    \includegraphics[width=0.45\textwidth]{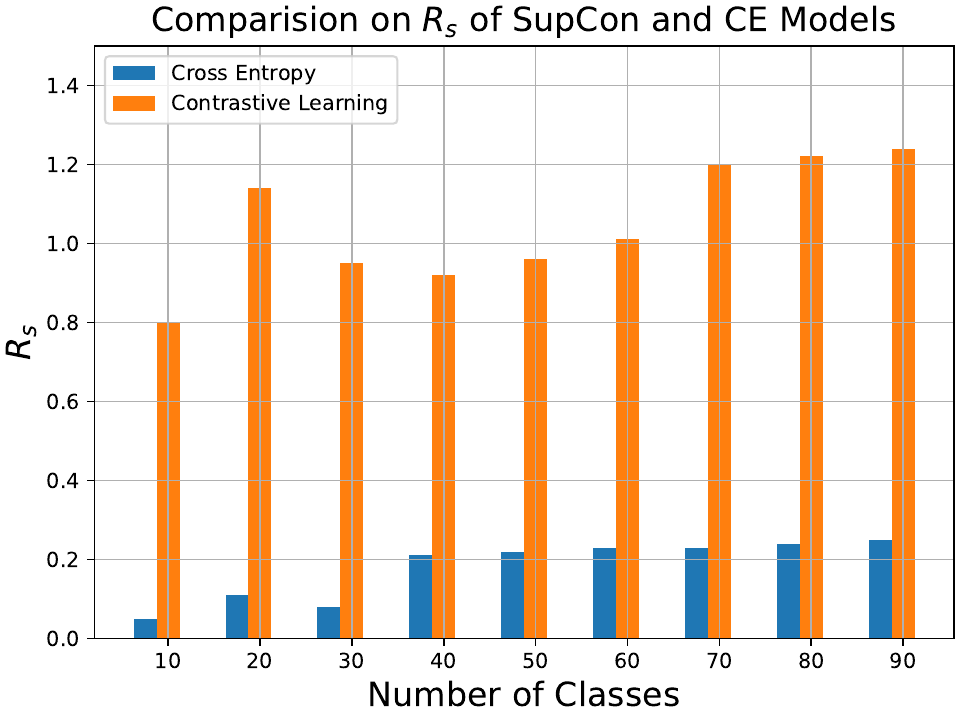}
  \end{center}
  \caption{$R_s$ of the feature maps extracted from \emph{ResNet-18} trained on \emph{Cifar-100} using supervised contrastive loss and cross entropy loss respectively.}
  \label{fig-contra-ce}
\end{figure}

In order to prove its superiority for satisfying Assumption \ref{assumption-intra} and \ref{assumption-inter} as cross entropy loss, we compare their $R_s$ through image recognition task on \emph{Cifar-100} dataset \cite{cifar}. It can be found from Fig. \ref{fig-contra-ce} that the $R_s$ in the models trained using supervised contrastive loss are much lower than those trained using cross entropy loss no matter how many classes are. 
Meanwhile, as reported in \cite{zhang2021unleashing}, the cross entropy loss has less capability in reducing intra-class feature scattering than contrastive learning that can increase $S_{intra}$. Considering all the above, we adopt supervised contrastive learning for our approach.

\subsection{Relation-based Knowledge Distillation} \label{relation-KD}

As introduced above, relation-based knowledge distillation is better for preserving feature relations. Inspired by \cite{park2019relational}, we utilize a temporal angle-wise and distance-wise distillation method in this paper to transfer the structural knowledge between the models before and after incremental training. 

The angle-wise similarity, $\psi_{A}$, is illustrated in Equ. \eqref{equ-angle-similarity}, in which $\mathbf{z}_i, \mathbf{z}_j, \mathbf{z}_k$ represent a triplet of features. Let $\psi_{A}(\mathbf{z}_i^{t-1}, \mathbf{z}_j^{t-1}, \mathbf{z}_k^{t-1})$ and $\psi_{A}(\mathbf{z}_i^{t}, \mathbf{z}_j^{t}, \mathbf{z}_k^{t})$ represent the angle-wise feature similarities given by the models in training session $t-1$ and $t$ respectively (see Equ. \eqref{equ-angle-similarity}), the temporal angle-wise distillation loss is in Equ. \eqref{equ-distill-loss}. $N$ stands for the number of exemplars (see \ref{subsubsec-examplar} for exemplar management). The angle-wise distillation loss is finally composed using the sum over the $L_2$ norms of the angle-wise similarity changes of all triples between the training sessions of all the exemplars. 

\begin{equation}
\begin{split}
    & \psi_{A}(\mathbf{z}_i, \mathbf{z}_j, \mathbf{z}_k) = \cos \angle \mathbf{z}_i \mathbf{z}_j \mathbf{z}_k = \langle \mathbf{e}^{ij}, \mathbf{e}^{kj} \rangle, \\
    & \mathbf{e}^{ij} = \frac{\mathbf{z}_i - \mathbf{z}_j}{{\|\mathbf{z}_i - \mathbf{z}_j\|}_2}, \mathbf{e}^{kj} = \frac{\mathbf{z}_k - \mathbf{z}_j}{{\|\mathbf{z}_k - \mathbf{z}_j\|}_2}
    \label{equ-angle-similarity}
\end{split}
\end{equation}

\begin{equation}
    \begin{split}
        \mathcal{L}_{dis-A} = \sum_{i, j, k \in N} {\|\psi_{A}(\mathbf{z}_i^{t-1}, \mathbf{z}_j^{t-1}, \mathbf{z}_k^{t-1}) - \psi_{A}(\mathbf{z}_i^{t}, \mathbf{z}_j^{t}, \mathbf{z}_k^{t})\|_2}
        \label{equ-distill-loss}
    \end{split}
\end{equation}

Besides, distance-wise distillation in Equ. \eqref{eq-loss-dis} is also applied to enforce the transfer of structural knowledge that aims to penalize distance chances between the corresponding sample pairs. $\psi_{D}$ is \emph{Euclidean} distance in our settings. 

\begin{equation}
    \mathcal{L}_{dis-D} = \sum_{i, j \in N} {\|\psi_{D}(\mathbf{z}_{i}^{t-1}, \mathbf{z}_{j}^{t-1}), \psi_{D}(\mathbf{z}_{i}^{t}, \mathbf{z}_{j}^{t}))\|_2}
    \label{eq-loss-dis}
\end{equation}

The distillation loss is the sum of $\mathcal{L}_{dis-A}$ and $\mathcal{L}_{dis-D}$ as in Equ. \eqref{eq-dis}. $\lambda_{dis}$ stands for a hyperparameter to balance $\mathcal{L}_{dis-A}$ and $\mathcal{L}_{dis-D}$.

\begin{equation}
    \mathcal{L}_{dis} = \mathcal{L}_{dis-A} + \lambda_{dis} \cdot \mathcal{L}_{dis-D}
    \label{eq-dis}
\end{equation}

\subsection{OpenIncrement Framework}
The incremental learning method proposed in this paper belongs to rehearsal-based approaches. Exemplars from the observed classes should be selected and stored. We name our method \emph{OpenIncrementNN} (\emph{OpenIncrement} Nearest Neighbor).

\subsubsection{Exemplar Management} \label{subsubsec-examplar}

In order to select exemplars that are representative for each class, we propose an \emph{Isometric Sampling} method as shown in Alg. \ref{alg-sampling}\footnote{All the pseudo-code semantically follows \emph{python} and \emph{Numpy} style in this paper}. When the new task $T$ comes, a portion of the exemplars of old classes are randomly removed if the memory is fixed.
For the features $\mathbf{F}_c$ belonging to the newly observed classes, their centers $\mathbf{\mu_c}$ are firstly derived.  For each feature point $\mathbf{z}_n$ in $\mathbf{F}_c$, their Euclidean distances to $\mathbf{z}_n$, $\mathfrak{E}(\mathbf{\mu}_c, \mathbf{z}_n)$, are computed and sorted. Then the samples are selected isometrically according to the distances to their class centers. 

\begin{algorithm}
\caption{Isometric Sampling for Selecting Exemplars}
\label{alg-sampling}
\begin{algorithmic}[1]
\STATE \textbf{Input}: Task $T$ with $C_{new}$ new classes and $C_{old}$ old classes, class centers of the old classes $\{\mathfrak{c}_1, \mathfrak{c}_1,..., \mathfrak{c}_{C_{old}}\}$,
samples in new classes $\mathbf{X}_{new}$ and their features $\mathbf{F}_{new}$, rehearsal memory size R, exemplar set for old classes $\mathbf{E} = [\mathbf{E}_1, \mathbf{E}_2,..., \mathbf{E}_{C_{old}}]$ and their features $\mathbf{F}_{old}$.
\STATE \textbf{Output}: Updated exemplars set $\mathbf{E}'$ for $C_{new}$ and $C_{old}$.
\newline
\STATE \textbf{Initialize}: $r = R /  C_{old}$, $r' = R / (C_{new} + C_{old})$, $\mathbf{E}' \leftarrow \emptyset$.
\newline
\newline
\textbf{Update exemplars for old classes for fixed memory:}
\FOR{$c = 1, 2,..., C_{old}$}
     \STATE $\mathbf{I} = random.sample(range(r), r')$
     \STATE $\mathbf{E}' \leftarrow \mathbf{E}' \cup \mathbf{E}_c[\mathbf{I}]$
\ENDFOR
\newline
\newline
\textbf{Select exemplars for new classes:}
\STATE  Isometric sampling the indices $\mathbf{i} = range(0, N, r')$
\FOR{$c = 1, 2,..., C_{new}$}
    \STATE Read features belonging to class $c$ with $N$ instances, $\mathbf{F}_c \subset \mathbf{F}_{new}$, $\mathbf{F}_c = [\mathbf{z}_1, \mathbf{z}_2,..., \mathbf{z}_N]$
    \STATE Class center $\mathfrak{\mu_c} = \frac{1}{N}\sum_{i \in N} \mathbf{z}_n$
    \STATE Initialize distance set $\mathbf{D} \leftarrow \emptyset$
    \FOR{$\mathbf{z}_n \in \mathbf{F}_c$}
        \STATE $\mathbf{D} \leftarrow \mathbf{D} \cup \mathfrak{E}(\mathfrak{\mu_c}, \mathbf{z}_n)$
    \ENDFOR
    \STATE Sort $\mathbf{D}$ in ascending order and get the indices $\mathbf{I} \leftarrow \mathbf{argsort}(\mathbf{D})$
    \STATE $\mathbf{I} \leftarrow \mathbf{I}[:,:,N//r']$
    \STATE $\mathbf{E}' \leftarrow \mathbf{E}' \cup \mathbf{X}_{new}[\mathbf{I}]$
\ENDFOR
\end{algorithmic}  
\end{algorithm}

\subsubsection{Unified Classification for In- and Outliers}    \label{subsubsec-unified-classification}
The testing phase consists of two parts, namely open set recognition and inlier classification as illustrated in Fig. \ref{fig-framework}. The set of features of the exemplars in each observed class $c$ extracted by the backbone, $\mathbf{Z}_{exem}^{C}=\{\mathbf{z}_{exem, 0}^{c}, \mathbf{z}_{exem, 1}^{c},..., \mathbf{z}_{exem, {N_c}}^{c} \}$, is first stored. For each testing sample feature $\mathbf{z}_{i}$, we search for its K-nearest neighbors, $\mathbf{Z}_{K}^{c, i}$, in $\mathbf{Z}_{exem}^{c}$ with cosine similarity, which is denoted using $S(\cdot)$. 
We use $\mathbf{S}_{K}^{c, i}$ to denote the set of K similarities between $\mathbf{z}_{i}$ and $\mathbf{Z}_{exem}^{c}$. The overall set of K similarities between $\mathbf{z}_{i}$ and exemplar sets of all classes is $\{ \mathbf{S}_{K}^{0, i}, \mathbf{S}_{K}^{1, i},..., \mathbf{S}_{K}^{C, i} \}$. The score, $sc_{osr}$, for discriminating outliers is given in Equ. \eqref{equ-similar-score}, in which each $\mathbf{S}_{K}^{c, i}$ is first normalized and its maximum is selected as $sc_{osr}$.

\begin{equation}
    sc_{osr} = \argmax_c \frac {\mathbf{S}_{K}^{c, i}} {\sum_c \mathbf{S}_{K}^{c, i}} 
    \label{equ-similar-score}
\end{equation}

\noindent 
$sc_{osr}$ is then compared with a pre-defined threshold $\tau_{osr}$. If $sc_{osr}$ is smaller than $\tau_{osr}$, it is an outlier, otherwise inlier. 

We train a neural classifier inputted by the backbone features, $\mathbf{Z}_{exem}^{C}$, for inlier classification. The classifier consists of one fully-connected layer. In order to prevent data imbalance between old and new classes, we use exemplars of the new classes instead of the full dataset to train the classifier.

\section{Experiments} \label{sec-experiments}
We test our method for incremental learning and open set recognition on image recognition tasks. 
The code is open source online \footnote{\url{https://github.com/gawainxu/OpenIncremen.git}}.

\subsection{Settings}

\subsubsection{Datasets} \label{subsubsec-datasets}
We test our method on two open source datasets for image recognition, namely \emph{Cifar-100} \cite{cifar} and \emph{Tiny ImageNet} \cite{ILSVRC15}.
\emph{Cifar-100} contains 100 classes of natural color images belonging to 20 sub-classes and there are 600 images in each class. \emph{Tiny ImageNet} is the subset of \emph{ImageNet} \cite{ILSVRC15} and consists of 200 classes of natural images.  We split the above two datasets into multiple non-overlapping sessions for different continual learning tasks and to act as inliers and outliers.
The number of classes in each session is shown in Tab. \ref{tab-class-split}. The classes in the last session of each dataset are outliers. Datasets are augmented before training. The details of data augmentation are in Appendix \ref{app-da}.

\begin{table}[]
    \centering
    \begin{tabular}{c|c}
    \toprule
        Datasets                &   $\sharp$ Classes per Session/$\sharp$ Sessions    \\
    \hline
        \emph{CIFAR-100}        &   10/10    \\
        \emph{Tiny ImageNet}    &   20/10     \\
    
    \bottomrule
    \end{tabular}
    \caption{Class split settings of each dataset for class-incremental learning}
    \label{tab-class-split}
\end{table}

\begin{table*}[t]
\begin{center}
\begin{tabular*}{0.9\textwidth}{c @{\extracolsep{\fill}} cccc}
    \toprule
     Methods &  \multicolumn{2}{c}{\textbf{CIFAR-100}} & \multicolumn{2}{c}{\textbf{Tiny-ImageNet}} \\
    \hline
    $M_{size}$ &  500 & 2000 & 2000 & 5000 \\
    \hline
    \emph{Joint}     &  \multicolumn{2}{c}{74.44}  & \multicolumn{2}{c}{53.55}  \\
    \emph{SI} \cite{zenke2017continual}         & \multicolumn{2}{c}{17.26}  &  \multicolumn{2}{c}{6.58}  \\
    \emph{EWC} \cite{kirkpatrick2017overcoming} & \multicolumn{2}{c}{23.1}  &  \multicolumn{2}{c}{7.58}  \\
    \emph{LwF} \cite{li2017learning}            & \multicolumn{2}{c}{16.22}  &  \multicolumn{2}{c}{8.46}  \\
    \emph{ER}  \cite{rolnick2019experience}     & 22.10 & 38.58 & 12.14    &  27.2  \\
    \emph{iCaRL} \cite{rebuffi2017icarl}        & 46.52 & 49.82 &   13.38  & 13.98 \\
    \emph{LUCIR} \cite{Hou_2019_CVPR}           &  40.59 &  41.73 &  14.97 &  17.61     \\
    \emph{Der} \cite{buzzega2020dark}           & 36.60 & 51.89 & 34.75 & 36.73 \\
    \hline       
    \emph{Supcon Full}          &  \multicolumn{2}{c}{75.46}   &     \multicolumn{2}{c}{55.34}       \\
    \emph{CE+ResKD}             & 18.63 & 41.91 &  12.3 &  14.5  \\
    \emph{CE+RKD}              &  32.57 & 44.37  &  14.22  & 15.23                                  \\
    \emph{DeepIncrement} (\textbf{ours})        & \textbf{49.95} & \textbf{63.73} &  \textbf{33.65} &  \textbf{37.76}    \\
   \bottomrule
\end{tabular*}
\end{center}
\caption{Average accuracy (in \%) for inlier classification of class-incremental learning. Some results are from \cite{buzzega2020dark} and \cite{boschini2022class}. Bold indicates the best results except \emph{Joint}.}
\label{tab-CIL}
\end{table*}

\begin{table*}[h]
\begin{center}
\begin{tabular*}{0.9\textwidth}{c @{\extracolsep{\fill}} cccc}
    \toprule
     Methods & \multicolumn{2}{c}{\textbf{CIFAR-100}} & \multicolumn{2}{c}{\textbf{Tiny-ImageNet}} \\
    \hline
    $M_{size}$ & 500 & 2000 & 2000 & 5000 \\
    \hline
    \emph{Supcon Full}          &  \multicolumn{2}{c}{74.51}   &     \multicolumn{2}{c}{62.54}      \\
    \emph{CE+ResKD}             & 55.05 & 60.64 & 50.23  & 51.57   \\
    \emph{CE+RKD}              & 56.55 & 59.7  & 50.98  &  51.58                                          \\
    \emph{DeepIncrement}    &   \textbf{63.09}  & \textbf{72.23} & \textbf{56.88}  & \textbf{59.39} \\
   \bottomrule
\end{tabular*}
\end{center}
\caption{AUROC for open set recognition on \emph{CIFAR-100} and \emph{Tiny-ImageNet} datasets. $M_{size}$ stands for the memory size. Bold indicates the best results except \emph{Joint}.}
\label{tab-AUROC}
\end{table*}

\subsubsection{Network Architecture}
For all experiments, we use \emph{ResNet-18} \cite{he2016deep} (residual neural network) as the backbone model. The last fully-connected layer that outputs class probability is replaced by two fully connected layers, which are the \emph{head} in \cite{khosla2020supervised}. \emph{ResNet-18} is a convolutional neural network and was originally used for image recognition. The output dimension of the last layer is 512. As introduced above, the neural classifier for inliers is one fully-connected layer.

\subsubsection{Training} \label{subsubsec-training}
We conduct all experiments under the class-incremental learning settings. Following the class split protocol introduced in \ref{subsubsec-datasets}, the models were trained in a sequential way, i.e., session after session. The backbones are trained using \emph{Adam} optimizer and the classifiers are trained with stochastic gradient descent (SGD) strategy.
The model hyperparameters were selected using \emph{grid search} and their search space and final configurations are shown in Appendix \ref{app-hp}. The baselines in \ref{subsubsec-baselines} are trained using the configurations in the original paper if they are available.

\subsubsection{Baselines} \label{subsubsec-baselines}
We select the state-of-the-art incremental learning methods as our baselines for inlier classification, namely \emph{SI}\cite{zenke2017continual}, \emph{EWC}\cite{kirkpatrick2017overcoming}, \emph{LwF}\cite{li2017learning}, \emph{ER}\cite{rolnick2019experience}, \emph{iCaRL}\cite{rebuffi2017icarl}, \emph{LUCIR}\cite{Hou_2019_CVPR}, and \emph{Der}\cite{buzzega2020dark}. As mentioned in \ref{subsec-incremental-learning}, \emph{SI}, \emph{EWC}, and \emph{LwF} are parameter-based methods. \emph{ER}, \emph{iCaRL}, \emph{LUCIR}, and \emph{Der} are replay-based methods. 

For open set recognition, since there is no existing work or benchmarks on the same topic in literature, we take the three settings listed in Tab. \ref{tab-baselines} as baselines.

\begin{table}[h]
    \centering
    \begin{tabular}{cc}
    \toprule 
        Settings & Details \\
    \hline
        \multirow{2}{*}{CE+ResKD} & \multirow{2}{*}{\makecell{Cross-entropy loss with response-based \\ knowledge 
                                                              distillation in \cite{hinton2015distilling}}}         \\
         &              \\
         \vspace{2mm}
        \multirow{2}{*}{CE+RKD} & \multirow{2}{*}{\makecell{Cross-entropy loss with RKD } }           \\
         &              \\
         \vspace{2mm}
         \multirow{2}{*}{Supcon Full} & \multirow{2}{*}{\makecell{Supervised contrastive learning trained \\ in joint 
                                                                  fashion} }           \\
         &              \\
         \bottomrule
    \end{tabular}
    \caption{Baselines for open set recognition}
    \label{tab-baselines}
\end{table}

\subsubsection{Evaluation Metrics}  \label{subsubsec-evaluation}
As in most works, we evaluate the incremental learning method for inlier classification using \emph{accuracy} defined in Equ. \ref{equ-accuracy-IL}. $C_{inliers}$ is the number of correctly classified inliers whereas $N_{inliers}$ is the total number of inlier testing samples. 

\begin{equation}
    A_{incremental} = \frac{C_{inliers}}{N_{inliers}}
    \label{equ-accuracy-IL}
\end{equation}

Since the approach for OSR in \ref{subsubsec-unified-classification} requires setting the threshold manually, a direct result comparison with different thresholds is not reasonable.
A threshold-independent metric, the \emph{area under the receiver operating characteristic} (AUROC) curve \cite{narkhede2018understanding} is taken as the metric to evaluate the OSR method. In the AUROC curve, the true positive rate is plotted against the false positive rate by varying the threshold. The higher AUROC is, the better the method can detect outliers (if outliers are assumed positive). When the AUROC value is 0.5, it is basically equivalent to tossing coins.

\subsection{Results}
\subsubsection{Class-incremental learning} \label{subsec-CIL}

The experiment results for class-incremental learning are shown in Tab. \ref{tab-CIL}. \emph{Joint} stands for normal offline training with all class samples using \emph{cross entropy} loss.
Our method surpasses most of the state-of-the-art baselines. 
Compared with the three baselines proposed by us, besides \emph{Supcon Full}, our method is much superior to the rest two. The results can also suggest that supervised contrastive learning is more robust in continual learning than cross entropy.

\subsubsection{Open Set Recognition}
The results of open set recognition are in Tab. \ref{tab-AUROC}.
Similar to the results of the inlier classification, our method shows better performance than the two baselines. It should be noted that we use the exemplars to detect outliers even for \emph{Supcon Full}. We have observed during conducting the experiments that the OSR performance also drops with the increase of training sessions, which is the same as for inlier classification (see the plots in Appendix \ref{app-extra}).

\section{Conclusion \& Future Work}



In this work, we presented a unified framework for class-incremental learning and open set recognition (OSR). Our method achieves state-of-the-art inlier classification accuracy and excels over baseline models in OSR, yet there remains room for improvement.

The field is in its infancy, prompting several pivotal questions for further exploration:

While catastrophic forgetting seems to induce feature map distortion post-continual learning, the specific nature of forgotten class-specific features causing this distortion warrants investigation.
It remains to be determined whether the features vital for OSR and inlier classification are congruent, and how continual learning impacts feature retention.
Enhancing other continual learning methodologies for outlier detection is essential. Concurrently, strategies to embed continual learning within established OSR techniques, ensuring retained efficiency, should be explored.
Despite the inherent synergy between open set recognition and continual learning, research intersections remain sparse. We urge a focused exploration in this domain, given its significance in evolving adaptive and autonomous machine learning systems.


{\small
\bibliographystyle{ieee_fullname}
\bibliography{egpaper_final}
}

\newpage
\newpage
\appendix

\section{Experiment Details}

\subsection{Backbone Networks}
\emph{ResNet-18} and multilayer perceptron are the backbone models in our work. The networks are separated into encoder and head and their architectures are shown in Tab. \ref{tab-resnet}. In original settings for supervised contrastive learning, the training is in two phases: 1) the encoder and head are trained together using supervised contrastive loss; 2) the head is removed and a downstream classifier is trained with the encoder outputs.

\begin{table*}[]
\begin{center}
\begin{tabular*}{0.8\textwidth}{c @{\extracolsep{\fill}} cc}
    \toprule
        \multicolumn{2}{c}{\textbf{Block Name}}                   &  \textbf{Configurations}   \\
         \hline
\multirow{30}{*}{\textbf{Encoder}}         & Conv1       &   $7\times7\times64$, stride 1, \emph{Conv 2D}     \\
         \vspace{2\baselineskip}\\
         &\multirow{5}{*}{Conv2\_x}              &   $1\times1\times64$, stride 1, \emph{Conv 2D}    \\
         &                                       &   \emph{Batch Normalization}            \\
         &                                          &   $3\times3\times64$, stride 2, \emph{Conv 2D}    \\
         &                                      &   \emph{Batch Normalization}             \\
         &                                      &   $1\times1\times64$, stride 1, \emph{Conv 2D}    \\
         \vspace{2\baselineskip}\\
         &\multirow{5}{*}{Conv3\_x}              &   $1\times1\times128$, stride 1, \emph{Conv 2D}    \\
         &                                      &   \emph{Batch Normalization}             \\
         &                                      &   $3\times3\times128$, stride 2, \emph{Conv 2D}    \\
         &                                      &   \emph{Batch Normalization}             \\
         &                                      &   $1\times1\times128$, stride 1, \emph{Conv 2D}    \\
         \vspace{2\baselineskip}\\
         &\multirow{5}{*}{Conv4\_x}              &   $1\times1\times256$, stride 1, \emph{Conv 2D}    \\
         &                                     &   \emph{Batch Normalization}            \\
         &                                     &   $3\times3\times256$, stride 2, \emph{Conv 2D}    \\
         &                                     &   \emph{Batch Normalization}             \\
         &                                     &   $1\times1\times245$, stride 1, \emph{Conv 2D}    \\
         \vspace{2\baselineskip}\\
         &\multirow{5}{*}{Conv5\_x}              &   $1\times1\times512$, stride 1, \emph{Conv 2D}    \\
         &                                     &   \emph{Batch Normalization}            \\
         &                                     &   $3\times3\times512$, stride 2, \emph{Conv 2D}    \\
         &                                     &   \emph{Batch Normalization}             \\
         &                                     &   $1\times1\times512$, stride 1, \emph{Conv 2D}    \\
         \vspace{2\baselineskip}\\
         &Average Pooling                        &  $1\times1$, Average Pooling               \\
         \hline
\multicolumn{2}{c}{\multirow{3}{*}{\textbf{Head}}}         &  \emph{Flatten}                    \\
         &                                  &   \emph{Linear}$(512, 512)$    \\
         &                                  &   \emph{Linear}$(512, 128)$             \\
         \bottomrule
\end{tabular*}
\end{center}
\caption{Network architecture of \emph{ResNet-18} trained in our paper. The encoder consists of five blocks and the head is one fully connected layer. $\ast$ stands for the dimension size after flattening. It differs between different input image sizes.}
\label{tab-resnet}
\end{table*}

\subsection{Data Augmentation}  \label{app-da}
For supervised contrastive learning, we applied the data augmentation approaches of horizontal and vertical \emph{flip}, \emph{color jitter} ($brightness = 0.4$, $contrast = 0.4$, $saturation = 0.4$, $hue = 0.1$), \emph{gray scaling} ($probability = 0.2$), and \emph{Gaussian blurring} ($kernel size = 9$) on the images in \emph{CIFAR100} and \emph{Tiny ImageNet}.

\subsection{Hyperparameters}   \label{app-hp}
The hyperparameters of our method are from two categories, namely the backbone networks and \emph{OpenIncrementNN}. For the backbone networks, the hyperparameters are learning rate ($lr$), epochs for the $t$th training session ($E_t$), loss balance ($\alpha$ and $\lambda_{dis}$), and temperature in supervised contrastive loss ($\tau$). Hyperparameter in \emph{OpenIncrementNN} is the number of neighboring samples ($K$).
The hyperparameters in backbone networks were searched using grid search. The hyperparameters in \emph{OpenIncrementNN} were determined using heuristics. 
Their search space and values are listed in Tab. \ref{tab-hyper}.

\begin{table*}[h]
\centering
\begin{tabular}{cccccc}
    \toprule
     & \multicolumn{3}{c}{\textbf{Backbone Networks}}  &  \multicolumn{2}{c}{\textbf{\emph{OpenIncrementNN}}}    \\[1pt]
     &  Parameter   & Value   &   Search Space          &  Parameter  & Value    \\
    \midrule [1pt]
    \multirow{5}{*}{\textbf{CIFAR100}}  & \hspace{0.5cm} $lr$    &  0.001 & [0.001, 0.005, 0.01] & \hspace{0.5cm}  $K$ & 10  \\
                                        & \hspace{0.5cm} $E_0$   & 100 & [100, 200, 300]   &  \hspace{0.5cm}    & \hspace{0.5cm}    \\
                                        & \hspace{0.5cm} $E_{t>0}$   & 200  & [100, 200, 300]  &  \hspace{0.5cm}   &   \hspace{0.5cm}    \\
                                        & \hspace{0.5cm} $\alpha$    &  0.2 & [0.05, 0.1, 0.2]  & \hspace{0.5cm}   &  \hspace{0.5cm}  \\
                                        & \hspace{0.5cm} $\lambda_{dis}$    &  0.5 & [0.2, 0.5, 0.8]  & \hspace{0.5cm}  &   \\
                                        & \hspace{0.5cm} $\tau$    &  0.05  & [0.01, 0.05, 0.1]  & \hspace{0.5cm}   &    \\
    \vspace{2\baselineskip}\\
    \multirow{5}{*}{\textbf{Tiny ImageNet}}  & \hspace{0.5cm} $lr$  &  0.001 & [[0.01, 0.005, 0.001]] & \hspace{0.5cm} $K$ &  20 \\
                                             & \hspace{0.5cm} $E_0$  &  100 & [100, 200, 300]  &  \hspace{0.5cm}   &  \hspace{0.5cm}      \\
                                             & \hspace{0.5cm} $E_{t>0}$  & 200  & [100, 200, 300]  &  \hspace{0.5cm}    &  \hspace{0.5cm}    \\
                                             & \hspace{0.5cm} $\alpha$   &  0.2 & [0.05, 0.1, 0.2]  & \hspace{0.5cm}  &   \\
                                              & \hspace{0.5cm} $\lambda_{dis}$    &  0.5 & [0.2, 0.5, 0.8]  & \hspace{0.5cm}   &   \\
                                             & \hspace{0.5cm} $\tau$    &  0.05 & [0.01, 0.05, 0.1]  & \hspace{0.5cm}   &   \\
   \bottomrule
\end{tabular}
\caption{Hyperparameters in the backbone networks and OpenIncrement framework that adopted in our experiments. The hyperparameters of the backbone networks are determined using grid search. The search space is given as well.}
\label{tab-hyper}
\end{table*}

\section{Extra Results}  \label{app-extra}

We have plotted the changes in inlier classification accuracy and AUROC for OSR on \textbf{CIFAR-100} dataset to see that they are positively correlated. We think such a correlation is due to the change in feature maps, i.e., the feature distortions. We leave the research on this problem in future work.

\begin{figure*}[htb]
    \centering 
\begin{subfigure}{0.37\textwidth}
  \includegraphics[width=\linewidth]{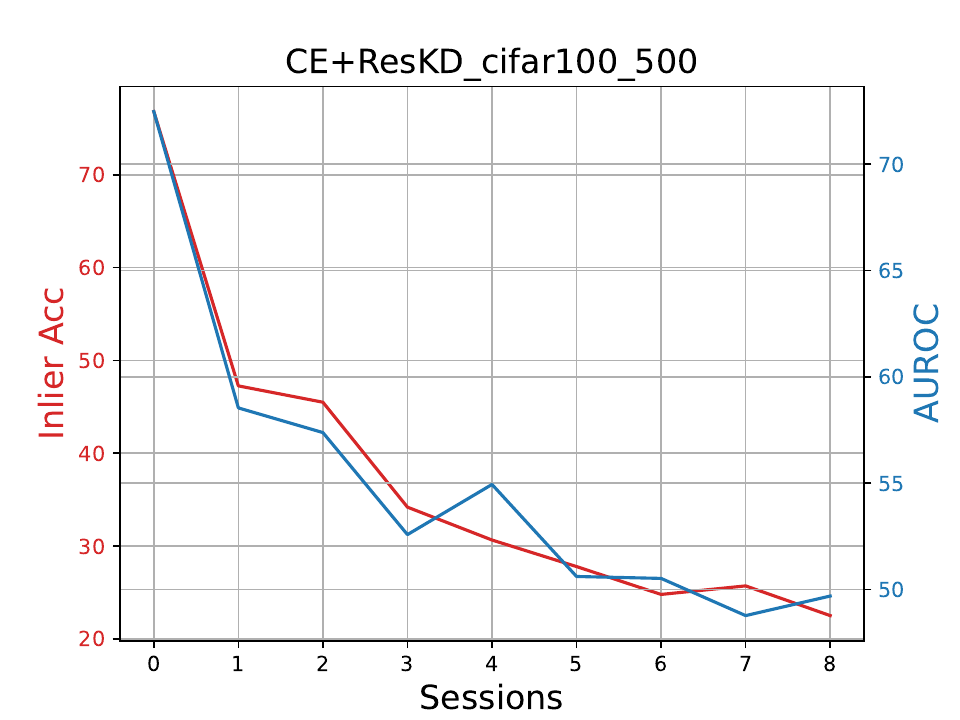}
  \caption{CE+ResKD $M_{size}$ 500}
  \label{fig:3a}
\end{subfigure}
\begin{subfigure}{0.37\textwidth}
  \includegraphics[width=\linewidth]{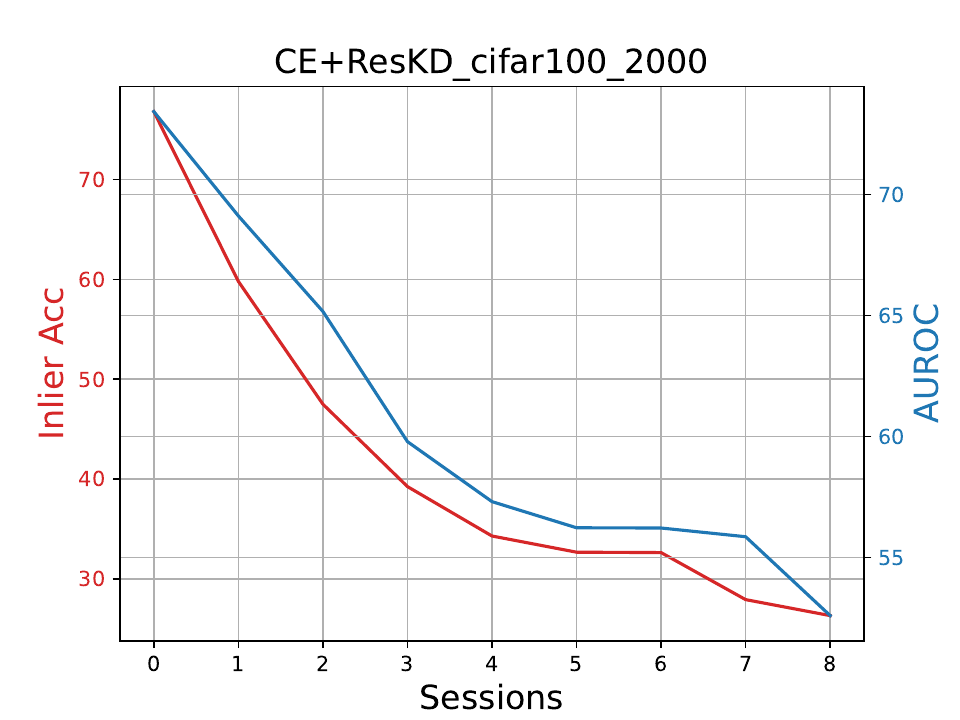}
  \caption{CE+ResKD $M_{size}$ 2000}
  \label{fig:3b}
\end{subfigure}
\\[-2.2mm]
\medskip
\begin{subfigure}{0.37\textwidth}
  \includegraphics[width=\linewidth]{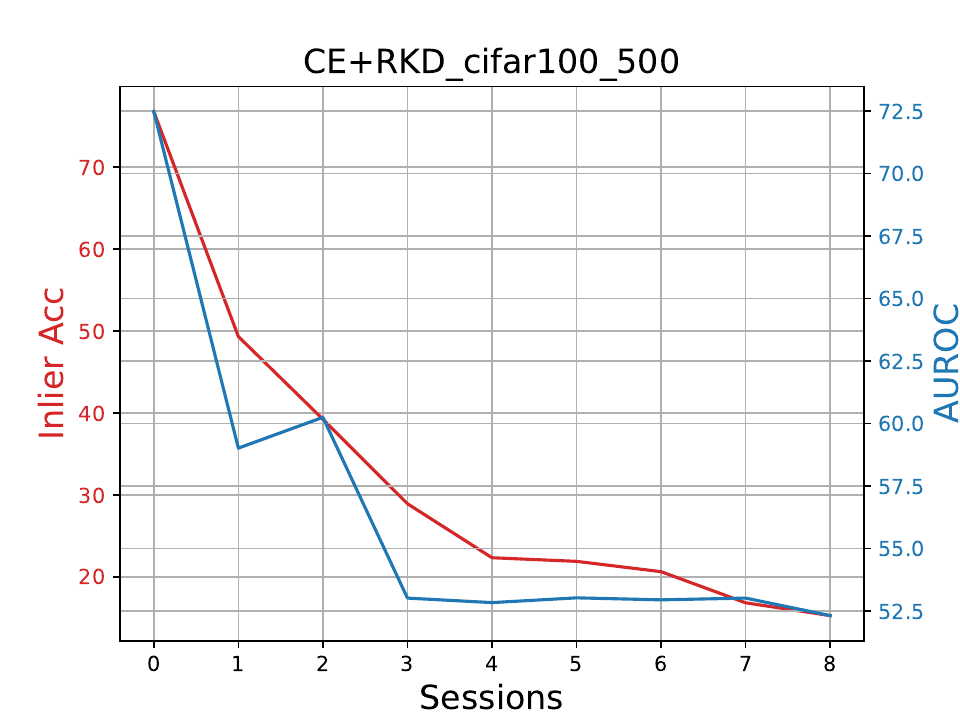}
  \caption{CE+RKD $M_{size}$ 500}
  \label{fig:3c}
\end{subfigure}
\begin{subfigure}{0.37\textwidth}
  \includegraphics[width=\linewidth]{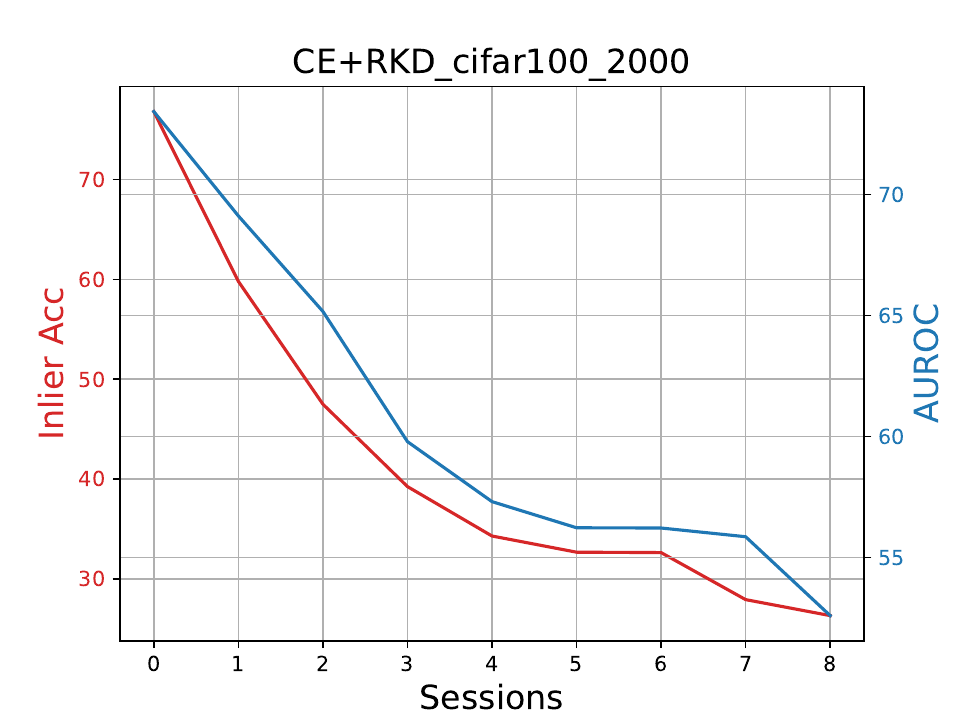}
  \caption{CE+RKD $M_{size}$ 2000}
  \label{fig:3d}
\end{subfigure}
\\[-2.2mm]
\medskip
\begin{subfigure}{0.37\textwidth}
  \includegraphics[width=\linewidth]{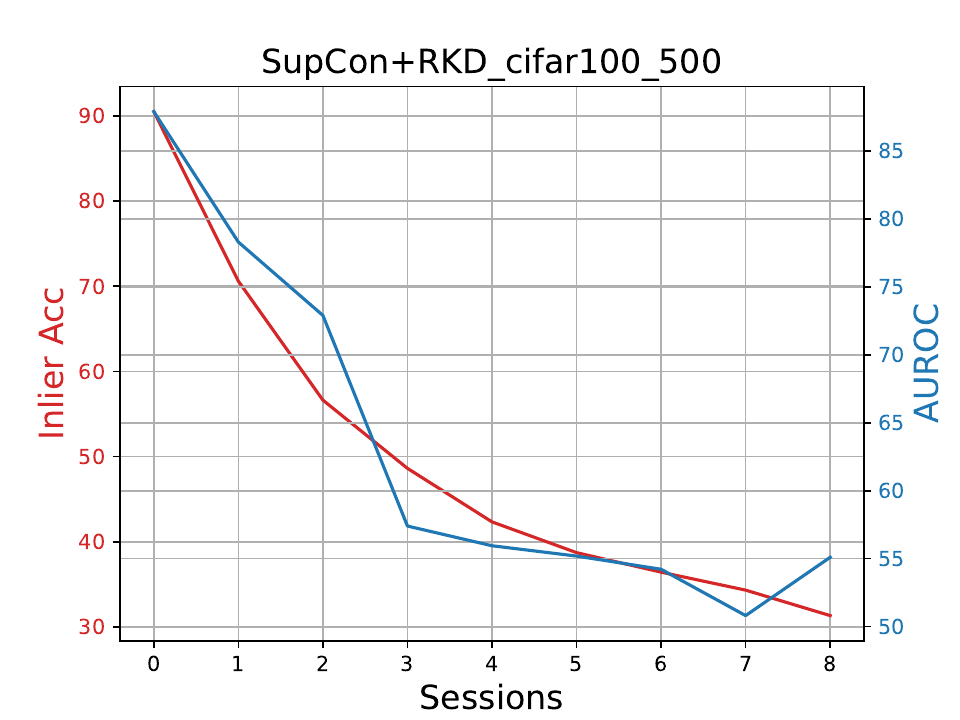}
  \caption{Supcon+RDK $M_{size}$ 500}
  \label{fig:3}
\end{subfigure}
\begin{subfigure}{0.37\textwidth}
  \includegraphics[width=\linewidth]{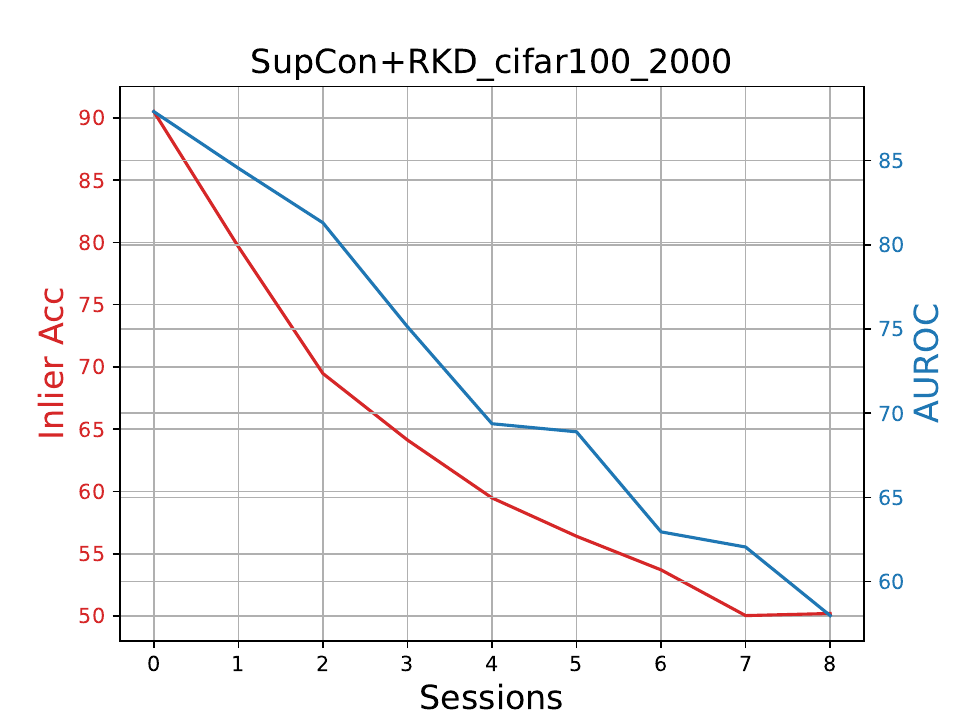}
  \caption{Supcon+RDK $M_{size}$ 2000}
  \label{fig:3f}
\end{subfigure}
\\[-2.2mm]
\medskip
\begin{subfigure}{0.37\textwidth}
  \includegraphics[width=\linewidth]{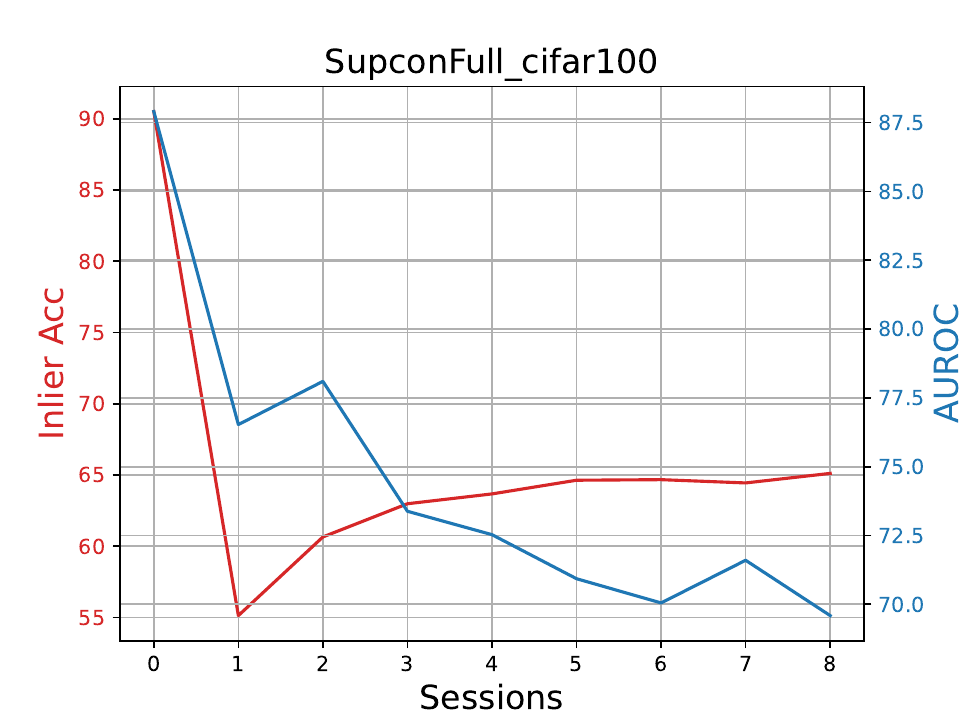}
  \caption{SupconFull}
  \label{fig:3g}
\end{subfigure}
\begin{subfigure}{0.37\textwidth}
  \includegraphics[width=\linewidth]{images/SupconFull_cifar100.pdf}
  \caption{SupconFull cifar100}
  \label{fig:3h}
\end{subfigure}
\caption{Joint plots on changes of inlier classification accuracy and AUROC for OSR on \textbf{CIFAR-100} dataset with the memory size of 500 and 2000. Since exemplar sizes in \emph{Joint} settings are all the same, the last two graphs are identical.}
\label{fig:images}
\end{figure*}

\end{document}